\documentclass[final,5p,times]{elsarticle}

\usepackage{booktabs}
\usepackage{tabularx}

\usepackage{amsmath}

\DeclareMathOperator*{\argmin}{arg\,min}
\DeclareMathOperator{\E}{\mathbb{E}}

\usepackage{hyperref}
\usepackage{xurl}

\usepackage{pgf-umlcd}
\usepackage{comment}

\usepackage{multirow}

\usepackage{amsmath}
\usepackage{amssymb}
\usepackage{float}
\usepackage{color}

\usepackage{algorithm}
\usepackage{algpseudocode}

\usepackage{multicol}
\usepackage{subcaption}

\usepackage{amsmath}
\usepackage{graphicx}
\usepackage{xcolor}
\usepackage{soul}
\usepackage[normalem]{ulem}

\graphicspath{ {./} }

\usepackage{caption}
\definecolor{cadetblue}{rgb}{0.37, 0.62, 0.63}
\definecolor{fuzzywuzzy}{rgb}{0.8, 0.4, 0.4}
\definecolor{babyblue}{rgb}{0.54, 0.81, 0.94}
\newcommand\norm[1]{\lVert#1\rVert}

\usepackage{tikz}
\usetikzlibrary{calc, matrix, positioning}
\usetikzlibrary{shapes, arrows, arrows.meta, shadows, positioning}

\tikzset{
  frame/.style={
    rectangle, draw,
    text centered,
    text width=6em, 
    minimum height=4em,fill=white,
    rounded corners,
  },
  initialize/.style={
    rectangle, draw,
    text centered,
    text width=16em, 
    minimum height=4em,fill=white,
    rounded corners,
  },
  block/.style={
    draw, rectangle, 
    text centered,
    fill=white,
    minimum height=3em, 
    minimum width=5em,
    rounded corners,
  },
  sum/.style={
    draw, circle,
    fill=white,
    node distance=1cm,
  },
  input/.style={coordinate},
  output/.style={coordinate},
  line/.style={
    draw, -{Latex},rounded corners=3mm,
  },
  pinstyle/.style={
    pin edge={to-,thin,black}
  }
}

\usepackage{subcaption} 

\usepackage[noabbrev]{cleveref}

\usepackage{amssymb}

\journal{Engineering Applications of Artificial Intelligence}

\begin{document}

\begin{frontmatter}

\title{Improved Long Short-Term Memory-based Wastewater Treatment Simulators for Deep Reinforcement Learning}

\author[inst1,inst2]{Esmaeel Mohammadi\corref{cor1}}
\ead{esm@kruger.dk}
\cortext[cor1]{Corresponding author.}
\author[inst3]{Daniel Ortiz-Arroyo}
\author[inst1]{Mikkel Stokholm-Bjerregaard}
\author[inst1]{Aviaja Anna Hansen}
\author[inst3]{Petar Durdevic}
\ead{pdl@energy.aau.dk}

\affiliation[inst1]{organization={Krüger A/S},
            addressline={Indkildevej 6C}, 
            city={Aalborg},
            postcode={9210}, 
            state={North Jutland},
            country={Denmark}}

\affiliation[inst2]{organization={Department of Chemistry and Bioscience, Aalborg University},
            addressline={Fredrik Bajers Vej 7H}, 
            city={Aalborg},
            postcode={9220}, 
            state={North Jutland},
            country={Denmark}}

\affiliation[inst3]{organization={AAU Energy, Aalborg University},
            addressline={Niels Bohrs vej 8}, 
            city={Esbjerg},
            postcode={6700}, 
            state={South Jutland},
            country={Denmark}}

\begin{abstract}
Even though Deep Reinforcement Learning (DRL) showed outstanding results in the fields of Robotics and Games, it is still challenging to implement it in the optimization of industrial processes like wastewater treatment. One of the challenges is the lack of a simulation environment that will represent the actual plant as accurately as possible to train DRL policies. Stochasticity and non-linearity of wastewater treatment data lead to unstable and incorrect predictions of models over long time horizons. One possible reason for the models' incorrect simulation behavior can be related to the issue of compounding error, which is the accumulation of errors throughout the simulation. The compounding error occurs because the model utilizes its predictions as inputs at each time step. The error between the actual data and the prediction accumulates as the simulation continues. We implemented two methods to improve the trained models for wastewater treatment data, which resulted in more accurate simulators: 1- Using the model's prediction data as input in the training step as a tool of correction, and 2- Change in the loss function to consider the long-term predicted shape (dynamics). The experimental results showed that implementing these methods can improve the behavior of simulators in terms of Dynamic Time Warping throughout a year up to \textbf{98}\% compared to the base model. These improvements demonstrate significant promise in creating simulators for biological processes that do not need pre-existing knowledge of the process but instead depend exclusively on time series data obtained from the system.
\end{abstract}


\begin{keyword}
Deep Reinforcement Learning \sep Dynamic Model \sep Simulator \sep Long Short-Term Memory \sep Phosphorus
\end{keyword}
\end{frontmatter}
\section{Introduction}
Process control is vital for wastewater treatment plants, as most industrial processes, since it will result in the desired outcome and reduction of costs \cite{zhang2012network}. Advanced control of wastewater treatment (WWT) processes to achieve proper effluent quality due to regulations with minimum costs and energy usage has been a significant topic in WWT research \cite{newhart2019data}. Model predictive control (MPC) is an advanced control method that utilizes a model of the system to predict future process parameters and manipulates the control variables based on that \cite{richalet1978model}. Even though MPC has shown better performance than conventional PID controllers with adaptive linear \cite{zhang2006adaptive} and nonlinear models \cite{bartman2009nonlinear} for WWT, there is still a need for better models and control approaches for most stochastic and nonlinear WWT processes \cite{hermansson2015model}. Other than MPC, artificial neural networks (ANN) have been used to predict the behavior of nonlinear dynamic systems to control the process parameters in WWT, such as Dissolved Oxygen (DO) and nutrient concentrations \cite{han2013nonlinear}. 

With the growth of artificial intelligence, different machine learning methods have been used to predict WWT parameters \cite{granata2017machine} and predictive control of the processes \cite{bernardelli2020real}. Reinforcement learning (RL) is a machine learning method that learns how to make sequential decisions through interactions with an environment. RL agents learn to maximize cumulative reward by exploring the environment, taking action, and observing the outcomes \cite{sutton2018reinforcement}. Although RL has been mainly used in the game-playing tasks \cite{silver2016mastering, vinyals2019grandmaster}, and robotics \cite{levine2016end, schulman2015trust}, there are some successful implementations of it in WWT process control \cite{syafiie2011model, chen2021optimal, hernandez2016energy}. In \cite{nian2020review}, some challenges of applying RL methods to industrial process control, such as data efficiency, stability, scalability, and transferability, have been discussed. A significant amount of progress has been made in developing one-step predictors for wastewater treatment, which forecast the next system step using only historical data \cite{hansen2022modeling, zhang2019predictive, kang2020time}. However, creating simulators for these systems remains an unaddressed research gap. In contrast to predictors, simulators start with historical data, but they become more self-sufficient as time goes on, using their predictions to make decisions \cite{mohammadi2024deep}. A simulator's capability to provide a virtual environment that enables agents to interact, learn, and optimize control policies is especially valuable in reinforcement learning \cite{sutton2018reinforcement}. 

When it comes to the implementation of RL in industrial processes, there are challenges posed by the lack of accurate simulators, some of which are explained in the following:
\begin{itemize}
    \item \textbf{Limited availability of high-fidelity simulators}: To model complex industrial processes, especially biological and chemical systems, accurately, detailed models must capture the complex dynamics and interactions within the system. However, the challenge is to develop high-fidelity simulators that accurately represent these processes among various operational conditions \cite{solle2017between}.
    \item \textbf{Model uncertainty and complexity}: There is often a high complexity and uncertainty associated with biological and chemical systems. Nonlinear dynamics, stochasticity, non-stationarity, parameter variations, and multiple interconnected subsystems should be considered when designing accurate simulators. Simulators cannot incorporate these complexities accurately and may introduce inaccuracies \cite{rasoulian2014uncertainty}.
    \item \textbf{Data shortage for training simulators}: Extensive experimental data is needed to calibrate and validate training simulators. Nevertheless, in many industrial processes, obtaining sufficient data is difficult due to limitations in data collection, experimental setups, and costs. Consequently, it is challenging to develop simulators that accurately reflect real-world dynamics \cite{qin2019advances, nian2020review}.
    \item \textbf{Transferability of models}: Models developed for one biological or chemical process may not readily generalize to other settings or variations. Achieving model transferability across different industrial environments and scenarios is a significant challenge, as each process may have unique characteristics and dynamics \cite{curreri2021rnn}. 
\end{itemize}

Additionally, it is often necessary to optimize single-step prediction loss to achieve accurate prediction for the next time step during forecasting tasks. This approach, however, may not be adequate when predicting over multiple time steps since errors may accumulate and result in suboptimal results. This paper explores alternative strategies that consider multi-step prediction losses and optimizes training methods to enhance the accuracy of multi-step simulations.

\section{Literature Review}
In \cite{venkatraman2015improving}, using forward simulation, authors present the "D\textsc{ata as} D\textsc{emonstrator}" method (D\textsc{a}D) for improving learned models. In D\textsc{a}D, prediction times and distributions are calculated after simulating the learned model's predictions on different scenarios. As a result of this data, a new model is trained, ensuring the predictions made at "test" time align with the correct future states. This method is similar to the Follow-The-Leader algorithm, involving the model being improved each time. The D\textsc{a}D method resembles interactive imitation learning, with training data as the "expert" who guides the correct course of action. As iteratively trained and simulated, the learned model demonstrates state-dependent action policies. By leveraging collected data, this method improves performance and enhances model predictions.

Traditional single-step prediction loss optimizes the model's ability to predict the immediate future accurately. Mathematically, the loss function can be expressed as follows \cite{mariet2019foundations}:
\begin{equation}
    \mathcal{L}_{single\_step} = \frac{1}{d_s} \displaystyle\sum_{d=1}^{d_s} {\norm{\hat{\mathbf{y}}_{d} - \mathbf{y}_{d}}^2}
\label{eq:single_step_loss}
\end{equation}

Where $d_s \in \mathbb{Z}^{+}$ is the number of the dimensions of the system's state ($s$, stating the system), while $\hat{\mathbf{y}}_{d} \in \mathbb{R}^{d_s}$ and $\mathbf{y}_{d} \in \mathbb{R}^{d_s}$ denote the predicted value and the corresponding ground truth at time step $t$ of the dimension $d \in \mathbb{Z}^{+}$ of the system. In contrast, multi-step prediction loss aims to improve the model's performance across multiple future time steps. The loss function is extended to incorporate predictions for $k \in \mathbb{Z}^{+}$ steps ahead \cite{mariet2019foundations}:
\begin{equation}
    \mathcal{L}_{multi\_step} = \frac{1}{k*d_s} \displaystyle\sum_{t=1}^{k}\displaystyle\sum_{d=1}^{d_s} {\norm{\hat{\mathbf{y}}_{t,d} - \mathbf{y}_{t,d}}^2}
\label{eq:multi_step_loss}
\end{equation}

Where $\hat{\mathbf{y}}_{t,d} \in \mathbb{R}^{k \times d_s}$ and $\mathbf{y}_{t,d}\in \mathbb{R}^{k \times d_s}$ denote the predicted and true values of the dimension $d$ of the system at time step $t$, respectively. 

DILATE (DIstortion Loss including shApe and TimE) was introduced in \cite{le2019shape} as a new objective function for training deep neural networks for multi-step and non-stationary time series forecasting. DILATE is designed to predict sudden changes in time series data accurately by incorporating two terms focusing on precise shape and temporal change detection. The DILATE loss function is expressed as:
\begin{equation}
\mathcal{L}_{\text{DILATE}}(\hat{\mathbf{y}}_i, \mathbf{y}_i) = \alpha \mathcal{L}_{\text{shape}}(\hat{\mathbf{y}}_i, \mathbf{y}_i) + (1 - \alpha) \mathcal{L}_{\text{temporal}}(\hat{\mathbf{y}}_i, \mathbf{y}_i)
\label{eq:dilate}
\end{equation}

Where $\hat{\mathbf{y}}_i \in \mathbb{R}^{k \times d_s}$ is the predicted trajectory, $\mathbf{y}_i \in \mathbb{R}^{k \times d_s}$ is the ground truth trajectory, and $\alpha \in [0,1]$ is a hyperparameter balancing the two terms. The shape term $\mathcal{L}_{\text{shape}}$ is based on a differentiable approximation of the Dynamic Time Warping (DTW) loss, which focuses on the structural dissimilarity between the predicted and actual signals. This term is defined as \cite{le2019shape}:
\begin{equation}
\mathcal{L}_{\text{shape}}(\hat{\mathbf{y}}_i, \mathbf{y}_i) = \text{DTW}_\gamma(\hat{\mathbf{y}}_i, \mathbf{y}_i) = -\gamma \log \sum_{\mathbf{A} \in \mathcal{A}_{k,k}} \exp\left(-\frac{1}{\gamma} \mathbf{A}, \Delta(\hat{\mathbf{y}}_i, \mathbf{y}_i)\right)
\label{eq:dilate_shape}
\end{equation}

Where $\mathcal{A}_{k,k}$ is the set of all valid warping paths in the binary matrix form connecting the endpoints (1,1) to (k,k). The warping path $\mathbf{A} \subset \{0,1\}^{k \times k}$ is a binary matrix that aligns the predicted time series $\hat{y}$ with the actual series $y$, indicating associations between their respective points \cite{rivest2019new}. The pairwise cost matrix $\Delta(\hat{\mathbf{y}}_i, \mathbf{y}_i)$ quantifies the dissimilarities, such as Euclidean distance, between each pair of predicted and actual values, measuring the alignment cost. The smoothing parameter $\gamma$, set at $10^{-2}$ in the study, plays a crucial role in DTW and Temporal Distortion Index (TDI) calculations, balancing the accuracy of shape matching with temporal flexibility in the time series forecasting model.

The Temporal Distortion Index is formulated to compute the temporal distortions between the predicted time series $\hat{\mathbf{y}_i}$ and the actual series $\mathbf{y_i}$. It is based on the deviation between the optimal Dynamic Time Warping path $\mathbf{A}^*$ and the first diagonal. The TDI loss function is mathematically represented as \cite{le2019shape}:
\begin{equation}
\text{TDI}(\hat{\mathbf{y}_i}, \mathbf{y_i}) = \langle \mathbf{A}^*, \Omega \rangle = \bigg\langle \underset{\mathbf{A} \in \mathcal{A}_{k,k}}{\argmin} \langle \mathbf{A},\Delta(\hat{\mathbf{y}_i}, \mathbf{y_i}) \rangle, \Omega \bigg\rangle
\end{equation}

Where $\Omega $ is a square matrix of size $k \times k$ that penalizes each element $\hat{\mathbf{y}}_{ih}$ being associated to an $\mathbf{y}_{ji}$ for $h \neq j$. The temporal loss term, $\mathcal{L}_{\text{temporal}}$, is defined to penalize temporal distortions between the predicted time series $\hat{y}_i$ and the actual series $y_i$. The equation for $\mathcal{L}_{\text{temporal}}$ is given as \cite{le2019shape}:
\begin{equation}
\mathcal{L}_{\text{temporal}}(\hat{\mathbf{y}}_i, \mathbf{y}_i) := \langle \mathbf{A^*}_\gamma, \Omega \rangle = \frac{1}{Z} \sum_{\mathbf{A} \in \mathcal{A}_{k,k}} \langle \mathbf{A}, \Omega \rangle \exp^{-\frac{\langle \mathbf{A}, \Delta(\hat{\mathbf{y}}_i, \mathbf{y}_i) \rangle}{\gamma}}  
\end{equation}

Where $\mathbf{A^*}_\gamma$ is the smoothed approximation of the optimal warping path, and $Z$ is the partition function, defined as \cite{le2019shape}:
\begin{equation}
Z = \sum_{\mathbf{A} \in \mathcal{A}_{k,k}} \exp^{-\frac{\langle \mathbf{A}, \Delta(\hat{\mathbf{y}}_i, \mathbf{y}_i) \rangle}{\gamma}}  
\end{equation}

This formulation aims to minimize temporal misalignments between the predicted and actual time series. The paper also describes efficient implementations for both loss terms' forward and backward passes, with a time complexity of $\mathcal{O}(k^2)$. The DILATE loss function outperforms other functions in experiments on various non-stationary datasets \cite{le2019shape}.

\subsection{Contributions}
According to the existing research on the creation of data-driven models for multi-step forecasting, wherein the accuracy of each forecasted step informs subsequent predictions, this study's contributions are as follows:
\begin{itemize}
    \item A new, improved model based on single-step prediction using only the actual data that has transformed it into a simulator capable of generating system states over a specific time horizon. This is achieved by initially sampling a single sequence of actual data.
    \item Investigation of how varying training episode structures, including controlled and randomized approaches, impact improved models' adaptability and learning efficiency. The results showed that introducing randomness to iterative training can improve the models' simulation accuracy by up to 22\% in terms of average mean squared error. 
    \item Implementing the loss of shape alongside the temporal loss in the improvement step of the model and studying its impact on the simulator.
    \item Creation of simulators for wastewater treatment processes using only time-series data from SCADA systems. These simulators aim to closely mimic the treatment plants, serving as the real plants' representatives for training deep reinforcement learning algorithms.
\end{itemize}

\section{Methods}
This section provides an overview of the data and methods used in our research. Initially, it details the wastewater treatment plant data utilized in this study. Following this, the section offers a concise introduction to dynamic systems modeling using the Long Short-Term Memory (\emph{LSTM}) model, including how the base model is trained to function as a simulator for reinforcement learning. The final part of the section elaborates on the D\textsc{a}D method, outlining the steps carried out in this research. This includes the incorporation of randomness in iterative improvement and the processes implemented for enhancing the simulator model.

\subsection{The Plant and Dataset}
This study focuses on the data from Kolding Central WWTP in Agtrup, Denmark. The time-series dataset for the period of two years was collected through the $\text{Hubgrade}^{\text{TM}}$ Performance Plant system, designed by Krüger/Veolia \cite{hubgrade}. Data preprocessing played a crucial role in enhancing model performance. Initially, the raw wastewater was normalized using the Min-Max technique, scaling the features to a range of 0 to 1. Feature selection was guided by principal component and correlation analysis. Variables of the system that demonstrated the highest correlation with the target variable, Phosphate concentration, were selected as inputs for the model, in conjunction with the target variable itself and the action variable, Metal dosage. More information about the plant, dataset, and preprocessing are provided in \cite{mohammadi2024deep}.

\subsection{Representation of dynamical systems by the \emph{LSTM} model}
Long Short-Term Memory (LSTM) models, first introduced by Hochreiter \& Schmidhuber in 1997 \cite{hochreiter1997long}, have become increasingly popular for time series forecasting due to their ability to efficiently handle complex, non-linear data and capture long-term dependencies \cite{hansen2022modeling, siami2018comparison}\cite{mohammadi2024deep}. These models are advantageous over others like ARIMA, as they can manage long-term dependencies, selectively remember or forget past information, handle variable-length input sequences, and be trained end-to-end \cite{siami2018comparison, yamak2019comparison}. Empirical studies show that LSTMs significantly outperform ARIMA models, reducing error rates by 84 - 87\%, and are robust as training iterations do not significantly affect performance \cite{siami2018comparison}. Additionally, \emph{LSTM} and Recurrent Neural Network (RNN) models are key in analyzing dynamic systems. They excel in capturing complex temporal dynamics and providing accurate predictions, especially for systems with fluctuating data patterns \cite{chen2022intelligent, gajamannage2023recurrent}. LSTMs are particularly adept at handling long-term dependencies, crucial in systems where stability is influenced by past events \cite{chen2022intelligent}. However, despite their versatility and effectiveness in modeling non-linear, real-world scenarios with time delays \cite{chen2022intelligent}, RNNs and LSTMs face challenges like being "black boxes" and requiring significant computational resources. Nonetheless, their adaptability and data-driven approach make them invaluable in dynamical systems analysis, promising avenues for future development \cite{chen2022intelligent, gajamannage2023recurrent, chang2019antisymmetricrnn}.

The \emph{LSTM}s and \emph{RNN}s can be utilized in control systems to model and predict dynamic processes over time. Let us consider a system with control input represented by a second-order linear differential equation:
\begin{equation}
    \frac{dx}{dt} = A\mathbf{x}(t) + B\mathbf{u}(t)
\label{eq:basic_diff}
\end{equation}

In this case, $\mathbf{x}(t) \in \mathbb{R}^{d_s}$ refers to the state vector at time $t$ (with $d_s \in \mathbb{Z}^{+}$ being the number of dimensions of the system), $\mathbf{u}(t) \in \mathbb{R}^{a_s}$ to the control input (with $a_s \in \mathbb{Z}^{+}$ being the number of control variables of the system), and $A$ and $B$ to the state and input matrices, respectively. Stability and controllability are analyzed using traditional control approaches, such as state-space representations and transfer functions.

Recurrent Neural Networks are used as one of the deep learning models to enhance the control framework. RNNs are ideal for modeling sequential data in dynamical systems with temporal dependencies. RNNs can update their state by solving the following equation:
\begin{equation}
    \hat{\mathbf{x}}(t+1) = \text{RNN}(A\mathbf{\mathbf{x}}(t) + B\mathbf{u}(t))  
\label{eq:RNN_state}
\end{equation}

As a result of RNN's hidden states, the system can retain memories of past states, enhancing its ability to predict the future. Additionally, we consider Long Short-Term Memory networks as an approach to solving the vanishing gradient problem while facilitating a long-term learning process. \emph{LSTM} networks can be used to update the state by calculating the following equation:
\begin{equation}
    \hat{\mathbf{x}}(t+1) = \text{LSTM}(A\mathbf{x}(t) + B\mathbf{u}(t))  
\label{eq:LSTM_state}
\end{equation}

For capturing long-term patterns, \emph{LSTM} cells use gating mechanisms to retain informed information. To use the \emph{LSTM} model for simulating the dynamical system, it should be able to start from a specific state of the system and try to act independently as the simulation goes on. This means that the trained \emph{LSTM} model will use its own predicted state of the system at each time step as input to predict the next state.

\subsection{Transition of the Model as a  Simulator for Reinforcement Learning}
In reinforcement learning, the state of the system at each time step is calculated by the simulation environment as follows \cite{mohammadi2024deep}:
\begin{equation}
{\hat{\mathbf{s}}_{t+1}} = f(\mathbf{s_t},a_t)
\label{eq:RL_state}
\end{equation}

Where $\hat{\mathbf{s}}_{t+1} \in \mathbb{R}^{d_s}$, $\mathbf{s}_{t+1} \in \mathbb{R}^{d_s}$, and $a_t \in \mathbb{R}^{a_s}$ are the predicted state of the system at time $t+1$, the state of the system, and the action taken from the agent at time $t$, respectively. At each step, $\mathbf{s}_t$ is obtained by the environment from equation \ref{eq:input_k}. Additionally, initializing from time step $t$ and running over $h$ episodes, the state of the system will be \cite{mohammadi2024deep}:
\begin{equation}
{\hat{\mathbf{s}}_{t+h}} = f(...f(f(\mathbf{s}_t,a_t),a_{t+1})...,a_{t+h})
\label{eq:states_normal}     
\end{equation}

If the prediction error at each time step is defined using the Euclidean distance, expressed as $e_t = \norm{\hat{\mathbf{s}}_t - \mathbf{s}_t}$, then the system's state following $h$ time steps can be described as \cite{lambert2022investigating, mohammadi2024deep}:
\begin{equation}
{\hat{\mathbf{s}}_{t+h}} = f(...f(f(\mathbf{s}_t,a_t)+e_t,a_{t+1})+e_{t+1}...,a_{t+h})+e_{t+h}
\label{states_error}     
\end{equation}

To develop a simulator that accurately forecasts the system's state after $h$ time steps, it's crucial to reduce the prediction errors $(e_t,e_{t+1},\dots,e_{t+h})$ throughout the simulation. A modified version of the D\textsc{a}D methodology, as outlined in \cite{venkatraman2015improving}, which incorporates variations in training to add randomness to trajectory distributions and durations, is employed in the current study. This approach enhances the model's ability to learn from and rectify its prediction errors at each step, thereby preventing the accumulation of errors.

\subsection{Training of the LSTM model}
The \emph{LSTM} architecture was designed with multiple layers to capture complex patterns in wastewater treatment data. Specifically, the network comprises two \emph{LSTM} layers, each consisting of 256 units, as described in \cite{mohammadi2024deep}. We employed the 'tanh' activation function in the \emph{LSTM} layers to facilitate non-linear learning. The model also included a dropout rate of 0.15 to prevent overfitting. The input to the model at each step consisted of a history of time steps, including all of the system's state variables, while the output was a single-step prediction of the system's state. The training and validation procedure of the base \emph{LSTM} model is explained in \cite{mohammadi2024deep}. 

The system's state and control variables at each time step $t$ are represented by $\mathbf{x}(t) \in \mathbb{R}^{d_s}$ and $\mathbf{u}(t) \in \mathbb{R}^{a_s}$, respectively. While the input provided to the \emph{LSTM} model consists of a historical record of the system's state $\mathbf{X}(t) \in \mathbb{R}^{l \times d_s}$ and control variables $\mathbf{U}(t) \in \mathbb{R}^{l \times a_s}$, detailed as follows:
\begin{equation}
    \mathbf{S}(t) = \mathbf{X}(t) + \mathbf{U}(t) = 
    \begin{bmatrix}
    \mathbf{x}(t)\\
    \mathbf{x}(t-1) \\
    \vdots \\
    \mathbf{x}(t-l)
    \end{bmatrix} + \begin{bmatrix}
    \mathbf{u}(t)\\
    \mathbf{u}(t-1) \\
    \vdots \\
    \mathbf{u}(t-l)
    \end{bmatrix}
\label{eq:input_t}
\end{equation}

Where $\mathbf{S}(t) \in \mathbb{R}^{l \times n}$ is the input to the \emph{LSTM} model at time $t$, with $n=d_s+a_s$ representing the number of the input variables. The output of the model at each time step $t$ will be as follows:
\begin{equation}
    \hat{\mathbf{S}}(t+1) =
    \begin{bmatrix}
    \hat{\mathbf{x}}(t+1)\\
    \hat{\mathbf{x}}(t+2) \\
    \vdots \\
    \hat{\mathbf{x}}(t+p)
    \end{bmatrix}
\label{eq:output_k}
\end{equation}

Where $p \in \mathbb{Z}^{+}$ represents the model's output sequence length, which in this study is set to $1$, consequently leading to $\hat{\mathbf{S}}(t+1) = \hat{\mathbf{x}}(t+1)$. According to equation \ref{eq:single_step_loss}, the prediction error of \emph{LSTM} model at time $t+1$ can be calculated as:
\begin{equation}
    \mathcal{L}_{t+1} = \frac{1}{d_s} \displaystyle\sum_{d=1}^{d_s} {\norm{\hat{\mathbf{x}}_{t+1,d} - \mathbf{x}_{t+1,d}}^2}
\label{eq:lstm_error}
\end{equation}

Common Training methods, known as \emph{teacher-forcing} or supervised learning, utilize backpropagation and minimize the single-step loss function for each training batch. Each training batch $\mathbf{D}$ contains $z \in \mathbb{Z}^{+}$ number of input and output sets sampled from the dataset $\mathcal{D}$ where $\mathbf{D}_i = ((\mathbf{S}(t)_i, \mathbf{x}(t+1)_i)$ and the optimization is done as the following \cite{venkatraman2015improving}:
\begin{equation}
    \mathbf{A^*} = \underset{A}\argmin  \displaystyle\sum_{i=1}^{z}\displaystyle\sum_{d=1}^{d_s} {\norm{(\hat{\mathbf{x}}_{t+1,d})_i - (\mathbf{x}_{t+1,d})_i}^2}
\label{eq:batch_minimize}
\end{equation}

Where $(\hat{\mathbf{x}}_{t+1,d})_i$ and $(\mathbf{x}_{t+1,d})_i$ are the predicted and true value of the dimension $d$ of the system's state at time $t+1$ for the pair $i$ of the batch $\mathbf{D}$. 

\subsection{D\textsc{ata as} D\textsc{emonstrator}}
In \cite{venkatraman2015improving}, the authors prove that recursive multistep prediction error for a discrete-time dynamical system is bounded in complexity by an exponential for a certain time horizon $T$:
\begin{equation}
    \norm{\hat{f}(\hat{\mathbf{x}}(T)) - \mathbf{x}(T+1)} \in \mathcal{O}(\exp(T\log(L))\epsilon)
\label{bounded_multistep}
\end{equation}

In the above equation, $\hat{f}$ is the learned model for the dynamical system, which is assumed to be Lipshitz continuous with constant $L > 1$, and bounded single-step prediction error $\epsilon \in \mathbb{R}$. To overcome this exponential growth of the prediction error, they use the D\textsc{a}D algorithm to train the learned model $\hat{f}$ on its own predicted distribution of states, which performs better than the original model. The true loss of the original model trained on only the real data without its own prediction can be defined as \cite{venkatraman2015improving}:
\begin{equation}
    \epsilon_N = \underset{f \in F}{\min} \frac{1}{N} \displaystyle\sum_{i=1}^{N} {\E_{x \sim P_{f_i}}[l_f(\mathbf{x})]}
\end{equation}

Where $\epsilon_N \in \mathbb{R}$ is the actual loss of the model, with a loss function $l$ and $F$ representing the class of models that includes a range of models suitable for regression tasks in the context of time series prediction. The distribution of the system's actual states and the model's prediction $(\mathbf{x}(t), \hat{\mathbf{x}}(t))$ is denoted as $P_f$, and $N \in \mathbb{Z}^{+}$ is the number of iterations, dictating how many times the learning and prediction processes are repeated. By considering the model returned by D\textsc{a}D as: 
\begin{equation}
   \hat{f} = \underset{f \in f_{1:N}}\argmin \E_{x \sim P_f}[l_f(\mathbf{x})]  
\end{equation}
As the number of iterations $N$ increases, the model identified by the D\textsc{a}D algorithm will have a performance that closely approaches the best possible model identified in hindsight \cite{venkatraman2015improving}:
\begin{equation}
    \E_{x \sim P_{\hat{f}}}[l_{\hat{f}}(\mathbf{x})] \leq \epsilon_N + o(1)
\label{eq:dad_final}
\end{equation}

Where the term $o(1)$ implies that as the number of iterations $N$ increases indefinitely, this particular term diminishes to zero, contributing less and less to the overall equation \cite{cormen2022introduction}.

\subsection{Iterative Improvement with D\textsc{a}D}
\label{sec:retrain_dad}
In the current study, the \emph{LSTM} model was first trained on the dataset of the real plant by dividing the training dataset into a specified number of batches and solving equation \ref{eq:batch_minimize} for each batch. Then, the already-trained model was optimized with a new training method, which helped to reduce prediction error in the simulation. First, the dataset was divided into several batches $\mathbf{D'}$; each including one pair of input  and outputs $\{\mathbf{S},\mathbf{Y}\}$, where $\mathbf{D'}_i = (\mathbf{S}(t)_i, (\mathbf{x}(t+1),...,\mathbf{x}(t+T))_i)$, and $T \in \mathbb{Z}^{+}$ is the episode length for each batch. For each input $\mathbf{S}(t)_i$, a dataset $\tilde{\mathcal{D}}$ containing the input and outputs of the model as a simulator was created as it is explained in Equation \ref{eq:input_k} and Figure \ref{fig:retrain_diagram}. The $\tilde{\mathcal{D}}$ dataset consists of input and output pairs as $\tilde{\mathcal{D}}=(\mathbf{S'}(t)_i, \hat{\mathbf{x}}(t+1)_i)$ and at each time step $k \in \mathbb{Z}^{+}$ where $0 < k < T$:

\begin{equation}
    \mathbf{S'}(k) = \mathbf{X}(k) + \mathbf{U}(k) =
    \begin{bmatrix}
    \mathbf{x}(k)\\
    \mathbf{x}(k-1) \\
    \vdots \\
    \mathbf{x}(k-l)
    \end{bmatrix} + \begin{bmatrix}
    \mathbf{u}(k)\\
    \mathbf{u}(k-1) \\
    \vdots \\
    \mathbf{u}(k-l)
    \end{bmatrix}
\label{eq:input_k}
\end{equation}

where:

\begin{align*}
\mathbf{x}(t) = \begin{cases} \mathbf{x}(t), & \text{if $t \leq 0$}.\\ \hat{\mathbf{x}}(t), & \text{otherwise}.\end{cases}
\end{align*}

While $\hat{\mathbf{x}}(t) \in \mathbb{R}^{d_s}$ is the output of the \emph{LSTM} model at time $t$, and $t \leq 0$ refers to the state of the system before the initialization point. The model's parameters are updated by solving the following for each batch:
\begin{equation}
    \mathbf{A^*} = \underset{A}\argmin  \displaystyle\sum_{j=0}^{T-1}\displaystyle\sum_{d=1}^{d_s} {\norm{\hat{\mathbf{x}}_{t+1+j,d} - \mathbf{x}_{t+1+j,d}}^2}
\label{eq:retrain_minimize}
\end{equation}

Resolving Equation \ref{eq:retrain_minimize} for each pair in $\mathbf{D'}$ results in a model, denoted as $\hat{f}$, which effectively reduces the prediction error across the entire simulation. This outcome is shown in Equation \ref{eq:dad_final} and has been proven in \cite{venkatraman2015improving}.

\newcommand\unitOne{1}
\tikzstyle{block} = [draw, fill=white, rectangle, minimum height=3em, minimum width=12em]
\tikzstyle{input} = [coordinate]
\tikzstyle{point} = [coordinate]
\tikzstyle{promtBlock} = [coordinate]
\tikzstyle{pinstyle} = [pin edge={to-,thin,black}]
\tikzstyle{sum} = [draw, fill=white, circle, node distance=1cm]

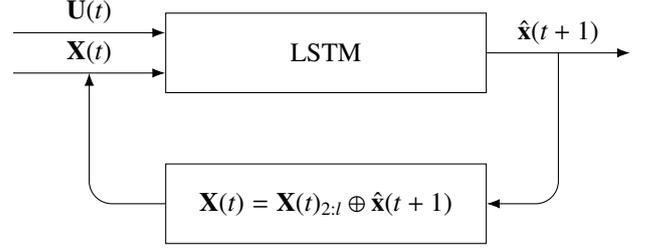
\begin{figure}
\centering
\begin{tikzpicture}[auto, node distance=2cm,>=latex']
    \node [block, align=center] (model) {LSTM};
    
    \coordinate (model-north) at ($(model.west)!.5!(model.north west)$)
    coordinate (model-south) at ($(model.west)!.5!(model.south west)$);
    
    \node [input, name=input_u, left=of model] at (model-north) {};
    \node [input, name=input_x, left=of model] at (model-south) {};
    \node [point, right of=input_x, align=center] (n1) {};
    \node [point, right of=model, align=center] (n2) {};
    \node [output, right of=n2, align=center] (output) {};
    \node [block, below of = model, align=center] (prediction) {$\mathbf{X}(t) = \mathbf{X}(t)_{2:l} \oplus \hat{\mathbf{x}}(t+1)$};
   
    \draw [line] (input_u) -- node [name=u] {$\mathbf{U}(t)$} (model-north);
    \draw [line] (input_x) -- node [name=x] {$\mathbf{X}(t)$} (model-south);
    \draw [line] (model) -- node [name=y] {$\hat{\mathbf{x}}(t+1)$} (output);
    \draw [line] (n2) (y) |- node {} (prediction);
    \draw [line] (prediction)  -| node [near end] {} (x);
\end{tikzpicture}
\caption{The diagram of the iterative dataset $\tilde{\mathcal{D}}$ aggregation process utilizing an \emph{LSTM} model, in which the model's predictions are fed back as input for the subsequent step. This routine is executed repeatedly for each input sequence's simulation horizon of $T$ iterations. The symbol $\oplus$ represents the concatenation operation used to append the predicted state of the system $\hat{\mathbf{x}}(t+1)$ as a new row to the historical states matrix $\mathbf{X}(t)$.}
\label{fig:retrain_diagram}
\end{figure}

\subsection{Randomness in Iterative Improvement}
Unlike \cite{venkatraman2015improving}, in this study, the prediction horizon $T$ was set to be not fixed in some of the experiments and instead was a random variable that could take on different values according to some probability distribution. Let $T \in \mathbb{Z}^{+}$ be a discrete random variable with a probability mass function $p(T)$, which gives the probability that the random variable $T$ equals $m$. The expected length of the prediction horizon is then:
\begin{equation}
  E[T] = \sum_{m=1}^{N} m \cdot p(m)
\end{equation}

Where $E[T]$ represents the expected value (mean) of the random variable $T$. The maximum value for the prediction horizon is stated by $N \in \mathbb{Z}^{+}$, and the summation is taken over all positive integers $m$. The learning objective needs to be redefined to minimize the expected prediction error over the random horizon $T$:
\begin{equation}
  \min_{f} E_{T \sim p(T)} \left[ \sum_{t=0}^{T-1} l\left(f(\mathbf{S}_t), \mathbf{x}_{t+1}\right) \right]
  \label{eq:random_learning}
\end{equation}
where $l$ is a loss function, $f$ is our model, and $\mathbf{S}_t$ is the history of the system's states at time $t$ as shown in Equation \ref{eq:input_k}. As stated in \cite{venkatraman2015improving} and demonstrated in Equation \ref{eq:dad_final}, the improved model $\hat{f}$, produced by the D\textsc{a}D method, guarantees better performance than the initial base model $f$, regardless of the length of the prediction horizon $T$.

\subsection{Improvement Procedure and Algorithm}
\label{sec:retrain_procedure}
The procedure of improving an already trained model with teacher-forcing and one-step prediction can be described as follows:

\begin{enumerate}
    \item Loading the previously trained model with the saved parameters, weights, and biases.
    \item Creating the $\mathbf{D'} \in \mathcal{D}$ with the actual values from the dataset, where the distribution of the input and output pairs will depend on the type of experiment. We define the following experiments to study the effect of randomness in the length of improvement for each pair and also the continuity of the batches:
    \begin{itemize}
        \item \textbf{E1 - Constant-Length, Consecutive Episodes:}
        This experiment serves as a baseline, using episodes of equal length in sequential order according to the dataset. The aim is to understand model performance under a stable and predictable training regime.
        \item \textbf{E2 - Random-Length, Consecutive Episodes:}
        While maintaining consecutive episodes, their lengths vary randomly. This introduces unpredictability in the amount of data processed in each episode, simulating realistic scenarios with fluctuating data input sizes.
        \item \textbf{E3 - Random Start, Constant-Length Episodes:}
        Episodes start at random points in the dataset but have a constant length. This tests the model's learning capabilities from data that is not sequentially organized, introducing randomness in data context while controlling quantity.
        \item \textbf{E4 - Random Episodes with Random Lengths:}
        Both the start points and lengths of episodes are random. This scenario tests the model's adaptability to an environment where both data context and quantity are highly variable, challenging its robustness and flexibility.
    \end{itemize}
    \item After the improvement for each batch is finished, the optimized model will be used for the next one.
    \item The process will be repeated for the specified number of iterations.
\end{enumerate}

The incorporation of randomness in training has been done for the following reasons:
\begin{itemize}
    \item \textit{Better Generalization:} Randomness in training can help prevent overfitting, improving generalization and performance on unseen data.
    \item \textit{Simulating Real-world Scenarios:} Training with randomness prepares the model for unpredictable and unstructured real-world data.
    \item \textit{Robustness and Flexibility:} Exposure to various levels of unpredictability enhances the model's robustness and flexibility, making it adaptable to changing or diverse data inputs.
\end{itemize}

In summary, these experiments systematically increase the level of randomness and unpredictability in training episodes to study how these factors influence the model's learning capabilities. This approach is crucial for developing models that are not only accurate but also adaptable and reliable in real-world, variable conditions.

The algorithm of the model improvement procedure is described in Algorithm \ref{alg:retrain}:
\begin{algorithm}[h]
\caption{The Model Improvement with Dataset Aggregation}
\label{retrain_code}
\begin{algorithmic}
\State \textbf{Inputs:}
    \Statex \hspace*{\algorithmicindent} \textit{Dataset $\mathcal{D}$, Model $f$, Type of the experiment, Epochs $N$,} \Statex \hspace*{\algorithmicindent} \textit{Min episode length, Max episode length}
\State \textbf{Outputs:}
    \Statex \hspace*{\algorithmicindent} \textit{The optimized model $\hat{f}$}
\State Initialization:
    \Statex \hspace*{\algorithmicindent} Build the parameters based on the type of the experiment
    \Statex \hspace*{\algorithmicindent} Build the input and output sets $\mathbf{D'}$ from $\mathcal{D}$ 
    \Statex \hspace*{\algorithmicindent} Set the optimized model parameters: $\hat{f} \gets f$
\For{epoch = 1, N}
    \For{$\mathbf{X}_{batch},\mathbf{Y}_{batch} \in \mathbf{D'}$}
    \State Initialize the prediction dataset $\tilde{\mathcal{D}}$
    \State $T \gets length(\mathbf{Y}_{batch})$
        \For{i = 1, $T$}
            \State $\mathbf{Y}_{pred,i} = f(\mathbf{X}_{batch})$
            \State Append $\mathbf{Y}_{pred,i}$ and $\mathbf{X}_{batch}$ to $\tilde{\mathcal{D}}$
            \State Update $\mathbf{X}_{batch}$ with $\mathbf{Y}_{pred,i}$
        \EndFor
        \State Aggregate all of the predictions from $\tilde{\mathcal{D}}$ to $\mathbf{Y}_{pred}$
        \State Calculate the loss $l$ with $\mathbf{Y}_{pred} \in \tilde{\mathcal{D}}$ and $\mathbf{Y}_{batch} \in \mathbf{D'}$ 
        \State Do backpropagation with the calculated loss $l$
        \State Update $\hat{f}$
    \EndFor
    \State Do a test simulation for $T \gets 1440$
    \State Calculate the simulation loss $l'$
    \State Save the current $\hat{f}$ if $l'$ is improved
\EndFor
\end{algorithmic}
\label{alg:retrain}
\end{algorithm}

\subsection{Software and Hardware}
All of the tests for the simulation environment are implemented in programming language \emph{Python} by using the \emph{Gym} \cite{brockman2016openai} and \emph{PyTorch} \cite{paszke2019pytorch} libraries. The AI Cloud service from Aalborg University is used for GPU-based computations. The used compute nodes are each equipped with 2 × 24-core \emph{Intel Xeon} CPUs, 1.5 TB of system RAM, and one \emph{NVIDIA Tesla V100} GPU with 32 GB of RAM, all connected via \emph{NVIDIA NVLink}.

\section{Results}
This section outlines the outcomes of the model enhancements and the methodology used for their computation. It includes a detailed presentation of metric calculations, the selection of optimal parameters for model improvement, and an analysis of the simulations conducted, all of which are discussed in the subsequent parts.

\subsection{Multi-Step Prediction Error Analysis}
\label{sec:multi_pred_error}
A similar approach to \cite{mohammadi2024deep} is done in this work to test the simulation environment. To do so, the pairs of actual input and output sequences $\{\mathbf{S},\mathbf{Y}\}$ were extracted from the dataset $\mathcal{D}$, where for the $i$th pair, ${\mathbf{S}(t)}_i$ and ${\mathbf{Y}(t)}_i$ are exactly as explained in Section \ref{sec:retrain_dad}. Each episode spans 1440 time steps, covering a full day's system state sequence. The Mean Squared Error (MSE) is computed as follows for the test sequences:
\begin{equation}
    \text{MSE} = \frac{1}{T \times d_s} \displaystyle\sum_{i=1}^{T}\displaystyle\sum_{d=1}^{d_s} {\norm{(\hat{\mathbf{x}}_{t+i,d}) - (\mathbf{x}_{t+i,d})}^2}
\label{eq:mse_test}
\end{equation}

Dynamic Time Warping, as explained in \cite{mohammadi2024deep}, was used to determine the similarities between the actual and predicted $\mathbf{Y}_i$ sequences for each pair of $\mathbf{S}_i$ and $\mathbf{Y}_i$. DTW is a method used to compare two time-series signals, like an actual signal and its predicted counterpart. It's particularly effective when these signals have similar patterns but may need to align better in time or speed \cite{cuturi2017soft}.

\subsection{The Improvement Parameters Selection}
The improved version of the base model was developed using the methods detailed in Section \ref{sec:retrain_procedure}. Several key parameters that influenced the enhanced model's performance were investigated:
\begin{itemize}
    \item \textbf{Minimum and Maximum Episode Length (Min EL, Max EL)}: These parameters define the episode lengths for each batch, with Min EL ranging from \textit{1} to \textit{10} and Max EL ranging from \textit{10} to \textit{2880}. These values remained consistent for Experiments 1 and 3.  
    \item \textbf{Loss Function}: We used either \textit{MSE} or \textit{DILATE} as the loss function for the model improvement. 
    \item \textbf{Alpha DILATE}: This parameter, ranging between \textit{0} and \textit{1}, was relevant when the DILATE loss function was employed.
\end{itemize}
The most influential improvement parameters, which resulted in the lowest average error in our simulations, are presented in Table \ref{tab:params_retrain}. For these simulations, 365 pairs of $\mathbf{S}$ and $\mathbf{Y}$ as explained in Section \ref{sec:multi_pred_error} were sampled from the dataset $\mathcal{D}$, representing an entire year. 
\begin{table*}[!hbt]
\centering
\caption{The parameters of the best improved checkpoints for each experiment}
\label{tab:params_retrain}
\begin{tabular}{@{}l|lllll@{}}
\toprule
\multicolumn{1}{c|}{\multirow{2}{*}{Experiment}}
& \multicolumn{5}{c}{Parameters} \\ \cmidrule(l){2-6} \multicolumn{1}{c|}{} &
Epochs & Min EL & Max EL & Loss Function & Alpha DILATE \\\midrule
    E1 &            50 &          1440 &          1440 &               Dilate &                 0.6 \\
    E2 &            50 &            10 &           480 &               Dilate &                 0.8 \\
    E3 &            50 &           240 &           240 &               Dilate &                   1 \\
    E4 &            50 &            10 &           480 &               Dilate &                   1 \\
\bottomrule
\end{tabular}
\end{table*}

\subsection{Simulation Analysis}
\begin{figure*}[!hbt]
    \resizebox{\textwidth}{!}{%
    \includegraphics[scale=1]{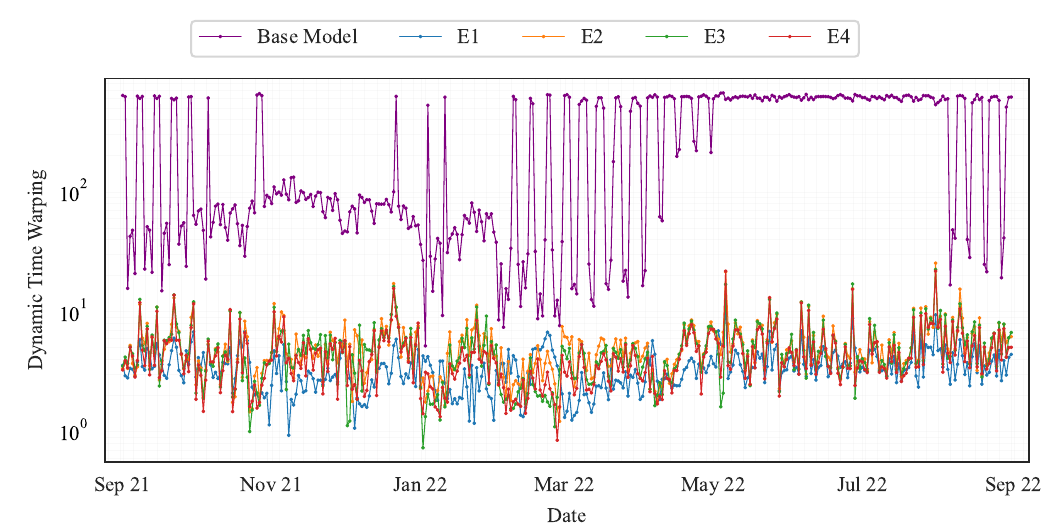}
    }%
    \centering
    \caption{Results of the simulation environment test for the base and improved \emph{LSTM} models.}
    \label{fig:mse_one_year}
\end{figure*}

\begin{figure*}
    \includegraphics[scale=1]{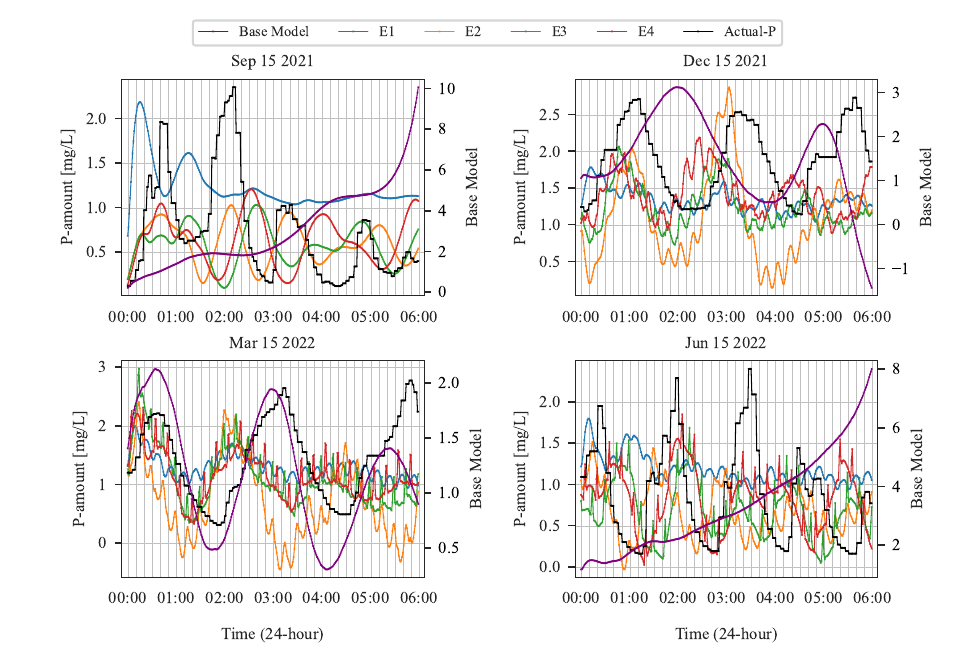}
    \centering
    \caption{Results of the simulation test for the four sequences of the dataset using the base and the improved \emph{LSTM} models}
    \label{fig:360_seasons}
\end{figure*}

The average MSE and DTW were computed for each optimized model. Figure \ref{fig:mse_one_year} illustrates the DTW associated with each test sequence of a whole year, where each experiment's improvement parameters were as stated in Table \ref{tab:params_retrain}.
The monthly average of MSE and DTW values for the experiments and Base Model are shown in Table \ref{tab:metrics_monthly}. Also, the simulation results for the four different sequences in the year, where the base and improved \emph{LSTM} models have been tested, can be seen in figure \ref{fig:360_seasons}, which the metrics are shown in Table \ref{tab:360_metrics_retrain}.
\begin{table*}
\centering
\caption{The average Mean Squared Error and Dynamic Time Warping data for the base model and improved versions during different months of the year. The best values of MSE and DTW for each month are highlightd in bold.}
\label{tab:metrics_monthly}
\begin{tabular}{@{}l|llllllllll@{}}
\toprule
\multicolumn{1}{c|}{\multirow{2}{*}{Month}}
& \multicolumn{2}{c}{Base Model} 
& \multicolumn{2}{c}{E1} 
& \multicolumn{2}{c}{E2} 
& \multicolumn{2}{c}{E3} 
& \multicolumn{2}{c}{E4} \\ \cmidrule(l){2-11} \multicolumn{1}{c|}{} &
MSE & DTW & MSE & DTW & MSE & DTW & MSE & DTW & MSE & DTW \\\midrule
     Sep &             0.8705 &            286.8591 &             0.8026 &    \textbf{3.9543} &             0.5438 &             5.8424 &    \textbf{0.5129} &              5.884 &             0.5579 &              5.455 \\
     Oct &             0.2938 &            134.7758 &             0.6578 &    \textbf{3.4257} &             0.3245 &             3.9728 &    \textbf{0.2576} &             3.5991 &             0.3126 &             3.5997 \\
     Nov & \underline{0.0746} &             89.9635 &             0.4795 &    \textbf{2.8723} &              0.735 &             5.2768 &             0.5725 &             5.0977 &             0.5715 &             4.5916 \\
     Dec &    \textbf{0.1221} &             87.9367 &             0.6485 &    \textbf{3.0582} &             0.9519 &             6.0672 &             0.7402 &             5.6731 &             0.7121 &             5.1503 \\
     Jan &    \textbf{0.1212} & \underline{77.9210} &             0.4855 &    \textbf{2.5644} &             0.5754 &             4.5247 &             0.4378 &             3.8361 &             0.3576 &             3.3124 \\
     Feb &             0.3581 &            145.7089 &             0.6184 &             3.7515 & \underline{0.2430} & \underline{3.2736} & \underline{0.1668} & \underline{2.2925} & \underline{0.1630} & \underline{2.2856} \\
     Mar &             0.9055 &            346.2727 & \underline{0.3672} & \underline{2.2502} &             0.5937 &             4.6793 &             0.4817 &             3.9503 &    \textbf{0.3654} &             3.3977 \\
     Apr &             1.3002 &            471.5401 &             0.5217 &    \textbf{3.1726} &             0.3567 &             4.6211 &             0.3963 &             4.3579 &    \textbf{0.3350} &             3.9808 \\
     May &             1.9043 &            613.0373 &             0.9235 &    \textbf{4.0940} &             0.6613 &              6.442 &             0.6472 &             5.8748 &    \textbf{0.6327} &             5.7668 \\
     Jun &             1.9415 &            614.4303 &              0.869 &    \textbf{4.0240} &             0.6584 &             5.4257 &    \textbf{0.5910} &             5.1915 &             0.6422 &             4.8706 \\
     Jul &              1.869 &             605.946 &             0.8044 &    \textbf{4.0203} &             0.8265 &             6.0527 &    \textbf{0.7475} &             5.8407 &             0.7521 &             5.3902 \\
     Aug &             1.3329 &            432.7499 &    \textbf{0.7785} &    \textbf{3.7837} &             0.8751 &             6.3217 &              0.818 &             6.0733 &             0.7989 &             5.4185 \\
 \midrule Average &             0.9245 &            325.5951 &              0.663 &    \textbf{3.4143} &             0.6121 &             5.2083 &             0.5308 &             4.8059 &    \textbf{0.5168} &             4.4349 \\
\bottomrule
\end{tabular}
\end{table*}

\begin{table*}
\centering
\caption{The average Mean Squared Error and Dynamic Time Warping data for the base model and improved versions during different seasons of the year. The best values of MSE and DTW for each month are highlighted in bold.}
\label{tab:360_metrics_retrain}
{\begin{tabular}{@{}l|llllllllll@{}}
\toprule
\multicolumn{1}{c|}{\multirow{2}{*}{Models}}
& \multicolumn{2}{c}{Base Model} 
& \multicolumn{2}{c}{E1} 
& \multicolumn{2}{c}{E2} 
& \multicolumn{2}{c}{E3} 
& \multicolumn{2}{c}{E4} \\ \cmidrule(l){2-11} \multicolumn{1}{c|}{} &
MSE & DTW & MSE & DTW & MSE & DTW & MSE & DTW & MSE & DTW \\\midrule
 Autumn &             10.748 &             60.235 &              0.621 &              11.113 & \underline{0.2420} &              7.561 & \underline{0.4930} &              8.155 &              0.466 &    \textbf{7.1040} \\
 Winter &              1.738 &             15.619 &    \textbf{0.5550} &               10.83 &              0.906 &             11.155 &              0.719 &             10.647 &              0.631 &    \textbf{6.4680} \\
 Spring & \underline{0.2610} & \underline{8.0650} &              0.601 & \underline{10.0220} &              1.869 &              12.22 &              0.842 &              8.349 &              0.684 &    \textbf{7.5470} \\
 Summer &              9.801 &             56.106 & \underline{0.4960} &               10.05 &    \textbf{0.3990} & \underline{5.6240} &               0.55 & \underline{5.3260} & \underline{0.4090} & \underline{4.9840} \\
\midrule Average &              5.637 &             35.006 &              0.568 &              10.504 &              0.854 &               9.14 &              0.651 &              8.119 &    \textbf{0.5470} &    \textbf{6.5260} \\
\bottomrule
\end{tabular}}
\end{table*}

\begin{figure*}[t]
\centering
  \begin{subfigure}[c][][s]{\textwidth}
    \centering
    \includegraphics[width=\textwidth]{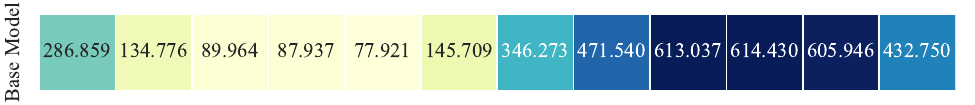}
    \caption{}
    \label{fig:heatmap_base}
  \end{subfigure}
  \\
  \begin{subfigure}[c][][s]{\textwidth}
    \centering
    \includegraphics[width=\textwidth]{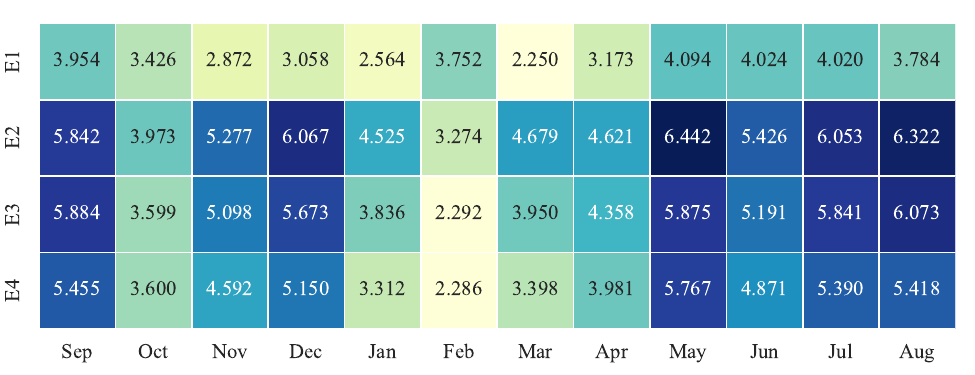}
    \caption{}
    \label{fig:heatmap_improved}
  \end{subfigure}
\caption{Heatmap of average dynamic time warping values per month for the base model (a) and the improved models (b).}
\label{fig:heatmap_mse_monthly}
\end{figure*}

\section{Discussion}
This study highlights the advancements in applying and improving the \emph{LSTM} networks for simulating nonlinear, dynamical systems. 

\subsection{Efficacy of Iterative Improvement Using Model's Predictions}
The innovative approach of improvement of the \emph{LSTM} model with its predictions (as discussed in Equation \ref{eq:retrain_minimize}) is crucial in enhancing the model's predictive accuracy and robustness. This methodological improvement is particularly effective in the following aspects:

\begin{itemize}
    \item \textbf{Adaptation to Dynamic Environments:} Wastewater treatment processes are highly dynamic, with varying conditions over time. By improvement of the \emph{LSTM} with its predictions, the model becomes more accurate at adapting to changing conditions.
    \item \textbf{Reduction of Overfitting:} Traditional training methods, which rely only on historical data, can lead to overfitting models and thus perform poorly on unseen data. The iterative improvement approach reduces this risk by continually challenging the model with new data points derived from its predictions.
\end{itemize}

\subsection{Advantages of the DILATE Loss Function}
Using the DILATE loss function (Equations \ref{eq:dilate} and \ref{eq:dilate_shape}) significantly contributes to the model's performance improvement. This loss function offers dual benefits:

\begin{itemize}
    \item \textbf{Enhanced Structural and Temporal Alignment:} By minimizing shape and temporal distortion, the DILATE loss function ensures that the \emph{LSTM} model accurately predicts wastewater treatment values and reproduces the timing and long-term dynamics of the wastewater treatment processes.
    \item \textbf{Improved Generalization:} The DILATE loss function aids in improving the model's generalization capabilities. Focusing on the predictions' structural integrity helps the model perform well on new, unseen data, which is essential for reliable simulations in real-world scenarios.
\end{itemize}

\subsection{Results and Performance Analysis}

\subsubsection{Comparison with Conventional \emph{LSTM} Models}
Comparative analysis demonstrates that the improved \emph{LSTM} model, guided by the DILATE loss function, outperforms conventional \emph{LSTM} models. This is evident in several key performance metrics, such as the mean squared error and dynamic time warping distance. The reduction in MSE, from the initial training to post-improvement, signifies the model's enhanced predictive accuracy:

\begin{equation}
    \Delta \text{MSE} = \text{MSE}_{initial} - \text{MSE}_{improved}
\end{equation}

Moreover, a lower DTW distance post-improvement indicates improved alignment with the temporal dynamics of the wastewater treatment process. According to the results shown in Table \ref{tab:metrics_monthly} and Figure \ref{fig:mse_one_year}, the accuracy of the base model to simulate the wastewater treatment plant data has been increased up to \textbf{98}\% in terms of dynamic time warping. These improvements are the results of training the model with its predictions fed as input at each time step, which will help the model improve its parameters based on the simulated input data. Also, the results of Figure \ref{fig:360_seasons} and Table \ref{tab:360_metrics_retrain} showed that our improved models exhibited an improvement of \textbf{90}\% (\textbf{E4}) in terms of MSE, and \textbf{81}\% (\textbf{E4}) in terms of DTW, compared to the base \emph{LSTM} model which has the exact same architecture as the model explained in \cite{mohammadi2024deep}. However, unlike \cite{mohammadi2024deep}, which used a traditional training method, our improved models benefit from the D\textsc{ata as} D\textsc{emonstrator} approach, enhancing their real-time applicability in dynamic environments.

\subsubsection{Validation Against Real-World Data}
Validation results using real-world data further confirm the model's efficacy. The model demonstrates high accuracy in predicting wastewater treatment dynamics and robustness in handling varying operational conditions. This robustness is quantified by examining the model's performance across different operational scenarios, reflecting its ability to generalize well beyond the training data.

\subsubsection{Detailed Analysis of Results}
A detailed monthly analysis revealed that the model's performance peaked in February, potentially due to the weather conditions or other disturbances, which led to the better capture of dynamics, as shown in Table \ref{tab:metrics_monthly} and Figure \ref{fig:mse_one_year}. The monthly analysis of results in this context does not support a broad conclusion that the model is more accurate in simulating the system's dynamics in certain months over others. Instead, the simulator's accuracy is primarily influenced by the system's real-time dynamics during simulation. This accuracy is subject to the impact of various system variables, disturbances, and potential operational failures or accidents.

\subsection{Broader Impact and Potential for Generalization}
This research contributes to environmental sustainability by enhancing the efficiency of wastewater treatment, thus supporting regulatory compliance and promoting cleaner water bodies. Additionally, given its success in this context, there is potential to generalize this \emph{LSTM} approach to other treatment processes, such as Ammonia removal, sludge treatment, and Nitrous oxide reduction, potentially revolutionizing process control in various industrial applications.

\section{Conclusions}
This research represents a significant advancement in \emph{LSTM} network applications for modeling dynamic systems, particularly in WWT simulations. The innovative iterative improvement approach using the model's predictions has notably improved the \emph{LSTM} model's accuracy and robustness. This method has shown its effectiveness in adapting to dynamic environments and reducing the risk of overfitting, making it suitable for complex environmental simulations. The employment of the DILATE loss function has played a critical role in improving the \emph{LSTM} model's performance. This loss function, which focuses on shape and timing alignment, makes it possible to predict wastewater treatment values accurately and reproduce the processes' timing and dynamics. In terms of performance, the redundant models showed an improvement of up to 90\% in Mean Squared Error and 81\% in Dynamic Time Warping compared to the base \emph{LSTM} model. 

The improvements in MSE and DTW values demonstrate the model's increased predictive accuracy and its potential to significantly improve the simulators for process control in various industrial applications. The model's effectiveness in adapting to dynamic environments and its robustness against overfitting are crucial for applications in complex environmental simulations.

For future work, exploring the model's applicability to other dynamic systems beyond wastewater treatment would be valuable. Investigating other loss functions or hybrid approaches that combine the strengths of DILATE with other strategies could further enhance model performance. Additionally, applying the model in real-time operational environments would provide practical insights and highlight areas for further improvement. Lastly, integrating this \emph{LSTM} approach with other machine learning techniques could be helpful for advanced predictive modeling in environmental and industrial applications.

\section{Acknowledgements}
The RecaP project has received funding from the European Union’s Horizon 2020 research and innovation programme under the Marie Skłodowska-Curie grant agreement No 956454. Disclaimer: This publication reflects only the author's view; the Research Executive Agency of the European Union is not responsible for any use that may be made of this information.




\bibliographystyle{elsarticle-num} 
\bibliography{cas-refs}






\end{document}